\documentclass{article}
\usepackage{amsmath}
\usepackage{graphicx}
\usepackage{tabularx}
\usepackage{changepage}
\usepackage{wrapfig}
\usepackage[final]{corl_2020} % Uncomment for the camera-ready ``final'' version.

\title{Addressing reward bias in Adversarial Imitation Learning with neutral reward functions}

\author{
    Rohit Jena \\
    The Robotics Institute \\
    Carnegie Mellon University \\
    \texttt{rjena@andrew.cmu.edu}
    \And
    Siddharth Agrawal \\
    The Robotics Institute \\
    Carnegie Mellon University \\
    \texttt{siddhara@andrew.cmu.edu}
    \And
    Katia Sycara \\
    The Robotics Institute \\
    Carnegie Mellon University \\
    \texttt{sycara@andrew.cmu.edu}
}

\begin{document}
\maketitle

%===============================================================================

\begin{abstract}
    Generative Adversarial Imitation Learning suffers from the fundamental problem of reward bias stemming from the choice of reward functions used in the algorithm.
    Different types of biases also affect different types of environments - which are broadly divided into survival and task-based environments.
    We provide a theoretical sketch of why existing reward functions would fail in imitation learning scenarios in task based environments with multiple terminal states.
    We also propose a new reward function for GAIL which outperforms existing GAIL methods on task based environments with single and multiple terminal states and effectively overcomes both survival and termination bias.
\end{abstract}

\keywords{Imitation Learning, Reward bias, Adversarial Learning} 
%===============================================================================

\section{Introduction}
Adversarial Imitation Learning(AIL) algorithms \cite{gail} \cite{airl} \cite{dac} \cite{kuefler2017imitating} have been shown to achieve state of the art performance on a variety of imitation learning tasks.
Performance of AIL algorithms have been primarily evaluated in two kinds of broad settings.
The first kind of setting can be characterised as survival based environments.
For example, environments in Mujoco like Hopper, Cheetah where the agent is rewarded to survive in the environment by remaining in a set of good states and penalized (by negative rewards or by termination) if it attains states other than these good states.
For example, in Hopper the agent gets good rewards for hopping forward as fast as possible.
If the agent gets into a set of `bad poses' the episode is terminated.
The other setting can be characterised as ``task'' based environments where the agent is rewarded to complete a task (reaching a goal state, interacting with an object in the environment, dodging an obstacle) and so on.
The episode terminates once the agent completes the task.
However, there is another important setting which hasn't drawn much attention in the Imitation Learning community.
The episode may also terminate when the agent has failed to perform the task in a task based scenario.
For example, when the agent may alter the environment such that completing the task becomes impossible, such as breaking a glass that the agent was supposed to pick up.
These are environments where the agent is rewarded to complete a task, i.e to reach a certain state in the shortest amount of time after which the episode is terminated, however, there are some additional bad terminal states, upon reaching which the episode terminates and the agent doesn't receive a reward.
Most of the community has worked on environments which are either purely survival based or purely task based.
However, to show that current reward functions are biased, we need to use more complex environments.
To that end, we use different environments of the Gym-Minigrid\cite{minigrid} package.
This package contains environments with single and multiple terminal states, which would be ideal to test if learning is hindered by the inherent biases in the reward function itself.

Our experiments show that existing methods\cite{gail}\cite{dac} often get stuck in a suboptimal policy in task based environments with multiple terminal states.
We provide a rough theoretical sketch of the underlying issues in these methods.
We also propose a reward function which is able to achieve comparable results in task based environments with single terminal state and performs significantly better than existing methods in task based environments with multiple terminal states.

\section{Related Work}
Early works in Imitation Learning based on Inverse Reinforcement Learning \cite{irl} focus on recovering a reward function from expert demonstrations which is optimised using Reinforcement Learning.
These methods are slow since they require solving the RL objective in the inner loop. 
From a practical standpoint, the intermediate step of recovering reward function is not really necessary when the goal is to only learn to imitate the expert.

Recent works in Adversarial Imitation Learning \cite{gail} \cite{airl} \cite{infogail} \cite{bayesiangail} \cite{goalgail} focus on directly recovering the expert policy without the need for recovering the exact reward function.
These methods alleviate the problem of distributional shift faced by some of the more traditional approaches like Behavior Cloning and its variants \cite{bco} \cite{dagger} \cite{rossbagnell} \cite{daume}.
GAIL based methods are also very sample efficient than behavior cloning in terms of number of expert trajectories required to recover the expert policy.
Although AIL methods have had a fair amount of success in various environments, recent work \cite{dac} has highlighted the implicit bias in the formulation of its reward function which results in degradation of performance in certain environments. 
The work in Discriminator-Actor-Critic \cite{dac} show how the reward function used in GAIL for training the agent is biased which leads to degradation of performance in certain environments.
In GAIL, the agent focuses on how to maximize its rewards from the discriminator, which may not necessarily translate to following the expert policy.
%The  problem was addressed in this work by learning a reward for terminal states, and they are able to achieve good performance in survival based environments in Mujoco.
In task-based environments, the problem of implicit bias due to zero rewards for terminal states encourages the agent to stay in the environment to collect more positive rewards, hence ignoring the task.
Their method worked on the survival based agents, as well as on task-based robotic environments.
However, there are no results on the method's performance in task based environments with multiple terminal states.
Our experiments show their formulation doesn't work well in such task based environments with multiple terminal states.
To the best of our knowledge, \cite{dac} is the only work which explicitly addresses reward bias and proposes a way to solve it.

\section{Background}
\subsection{Generative Adversarial Imitation Learning}
We make use of the GAIL\cite{gail} framework to learn the expert policy. In  GAIL framework, to learn a robust reward function, a discriminator(D) is trained which acts like a binary classifier to differentiate between transitions from the expert and from the trained policy. The policy is trained using an on-policy RL algorithm, and it is rewarded for confusing the discriminator.

\[ \max_\pi \min_{D} \Big[ E_{a\sim \pi(s)} [\log(D(s,a))] + E_{a \sim \pi_E(s)} [\log(1 - D(s,a))] + \lambda H(\pi) \Big] \tag{1} \label{eq:gail}\]

where $\pi$ is the imitation learning policy,  $H(\pi)$ is the entropy of the policy, $\pi_E$ is the expert policy and D(s, a) is the probability that the state-action pair is from the expert trajectory.
% The reward learnt by GAIL might not correspond to the true reward of the environment but it can be used for matching the occupancy measure of the expert as demonstrated in the previous works\cite{gail}. 
GAIL does not aim to recover the true reward function, but rather tries to match the state occupancy measure of the expert and the agent.
However, this does not guarantee that the expert trajectories are optimal with respect of the reward function output by the discriminator as we show in the later sections.

\subsection{Survival Based and Task Based Environments}
In survival based tasks, the agent gets a reward for staying in a set of "good states" and the episode terminates when the agent goes to a "bad state".
Within the good states, different rewards may be provided to encourage the agent to stay within certain good states more than others.
Some examples of these kinds of environments are the Mujoco environments like Hopper, Reacher etc.

In task-based environments, the agent gets a reward for successfully completing a task, characterised by agent reaching a certain state in the environment.
Additional rewards may be given for completing `subtasks' along the way.
The agent may also be penalized for simply existing in the environment.
This makes the agent complete the task as soon as possible to get maximum positive rewards from task completion and minimum survival penalty.

Additional complexities might arise, if additionally the environment also has some other terminal states, which don't lead to completion of the task but lead to the termination of the episode or the agent essentially dying.
In our work, we use 3 environments of this kind and show how existing approaches\cite{dac} \cite{gail} don't work on these environments and how our method tackles the biases.

\subsection{Reward bias in GAIL}
To train the imitation policy in the GAIL method, two kinds of discriminator reward are commonly used in literature \cite{gail} \cite{goalgail}:
\begin{enumerate}
\item \textbf{Positive Reward}: This reward is always positive and its value is $-\log(1 - D(s, a))$
\item \textbf{Negative Reward}: This reward is always negative and its value is $\log(D(s, a))$
\end{enumerate}
However, both the reward functions have implicit biases which may hinder learning. 
They are explained in the following subsections.
\begin{figure}
    \centering
    \includegraphics[width=0.32\textwidth]{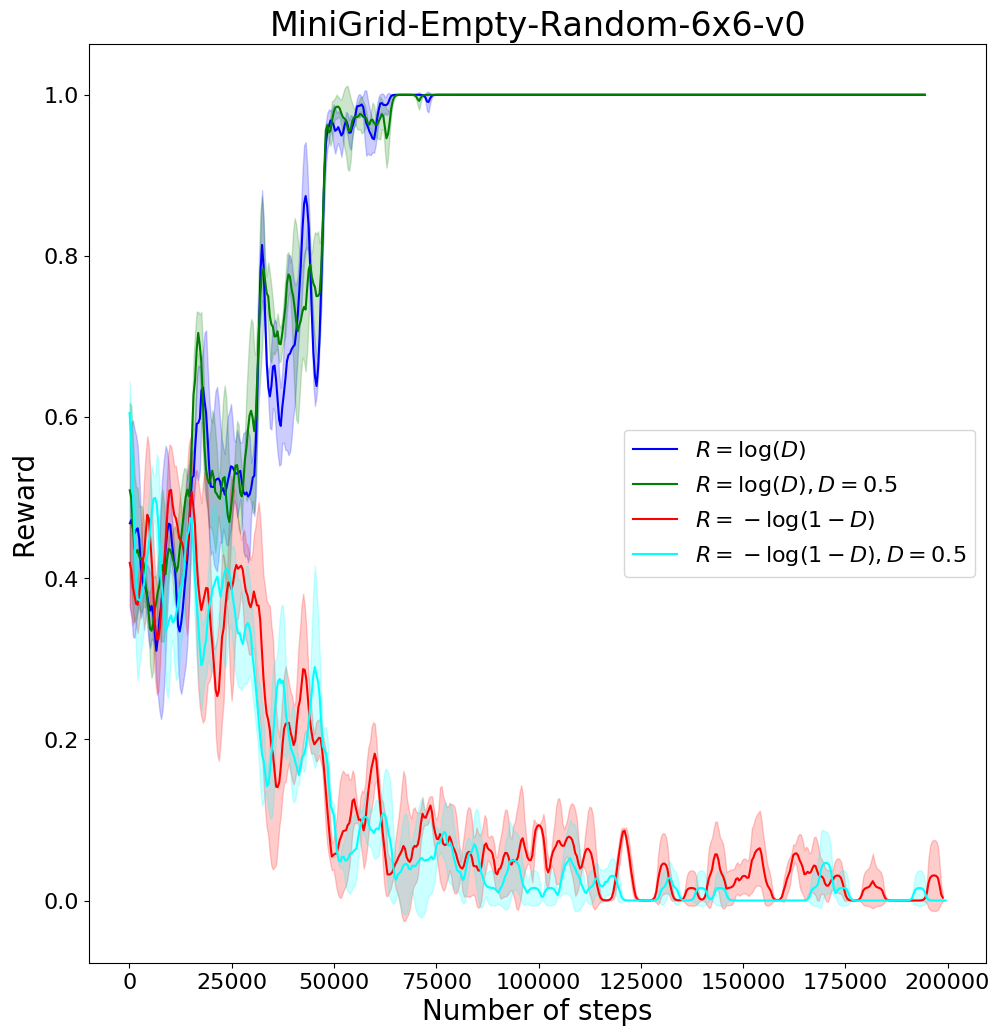}
    \includegraphics[width=0.32\textwidth]{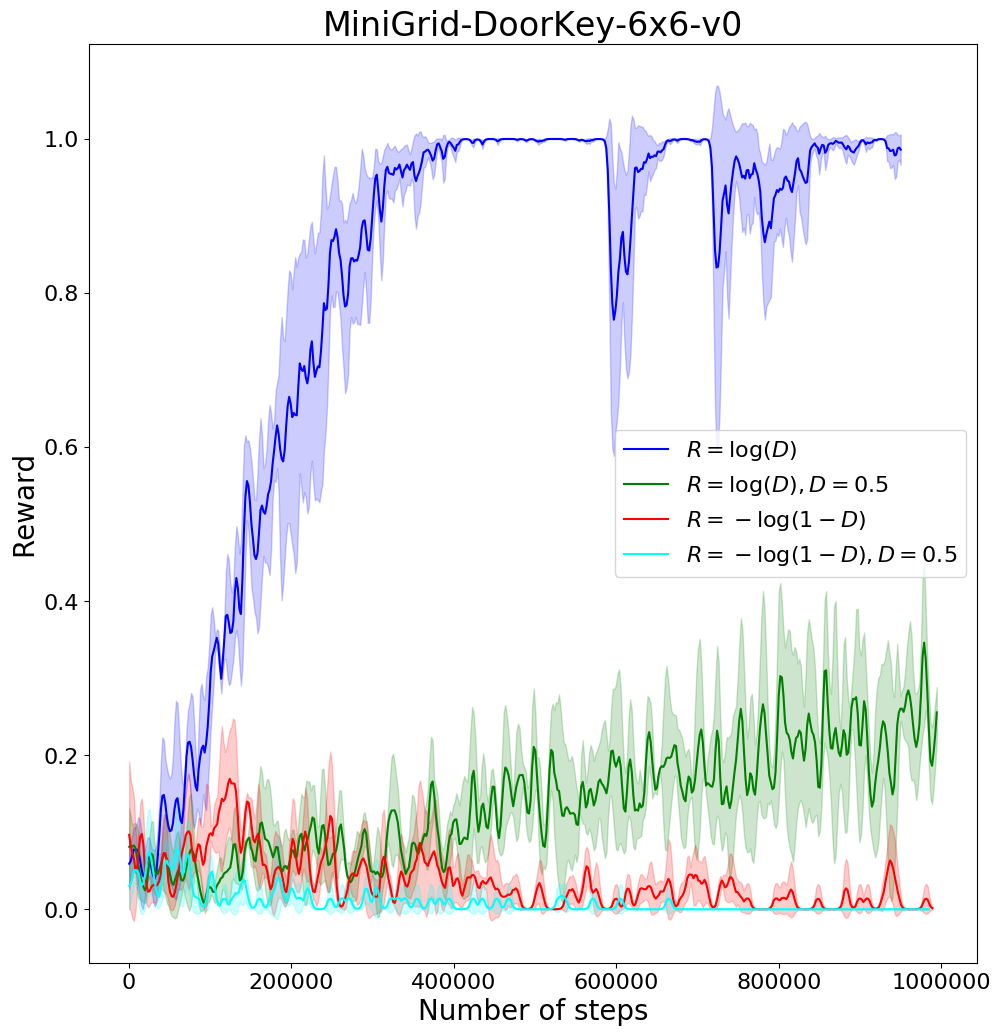}
    \includegraphics[width=0.32\textwidth]{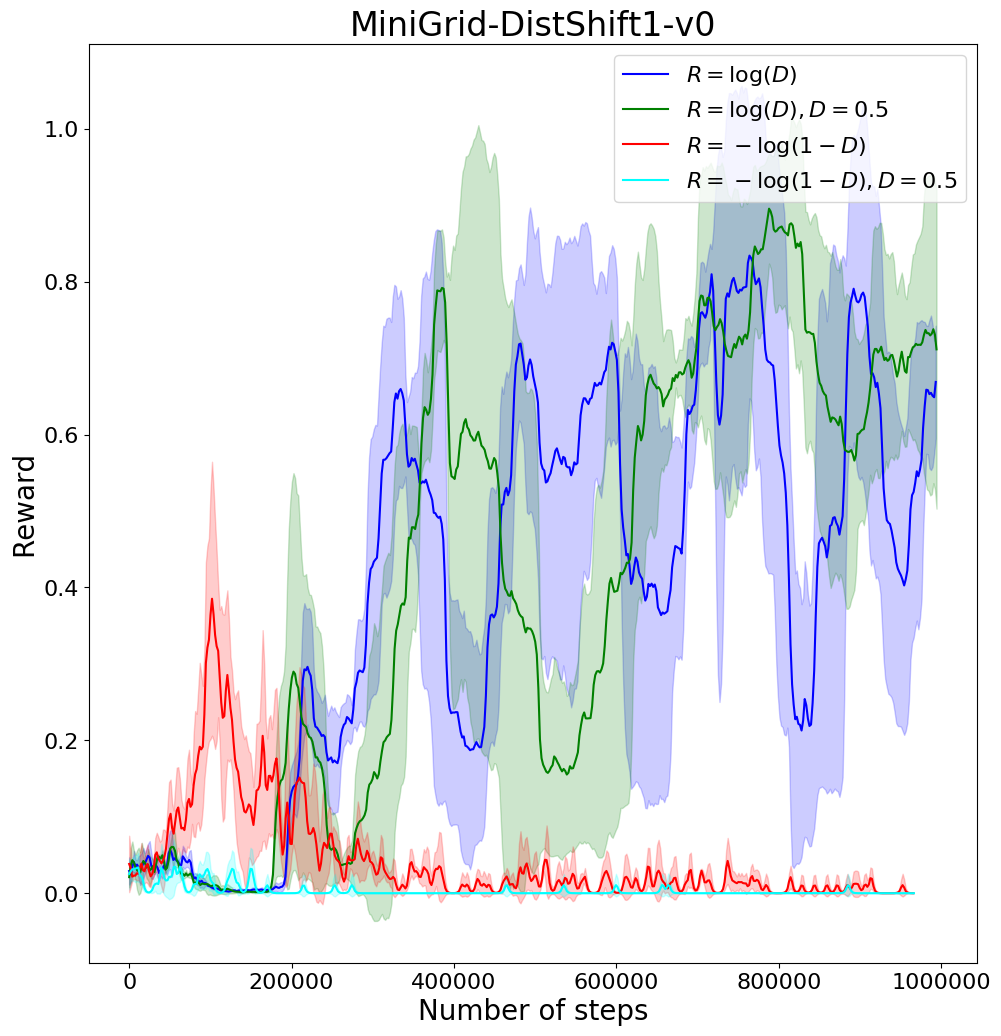}
    \includegraphics[width=0.32\textwidth]{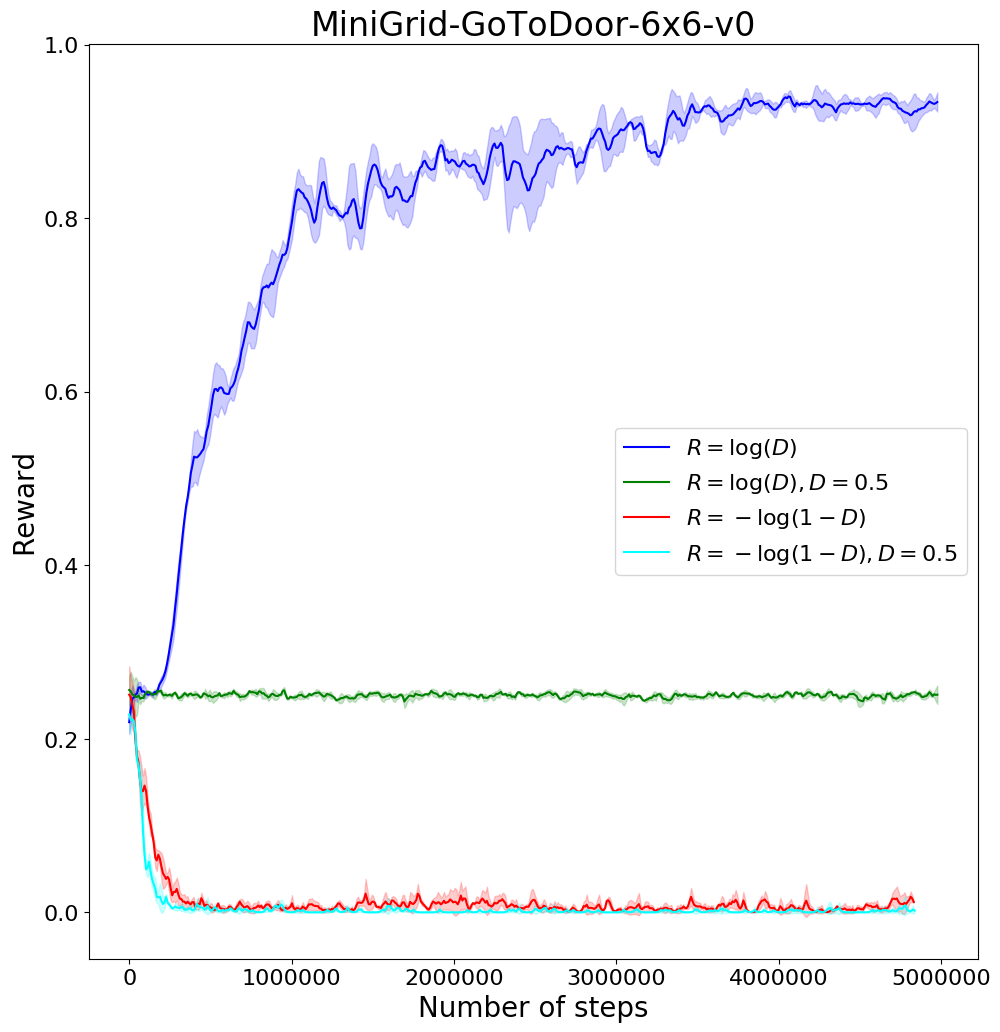}
    \includegraphics[width=0.32\textwidth]{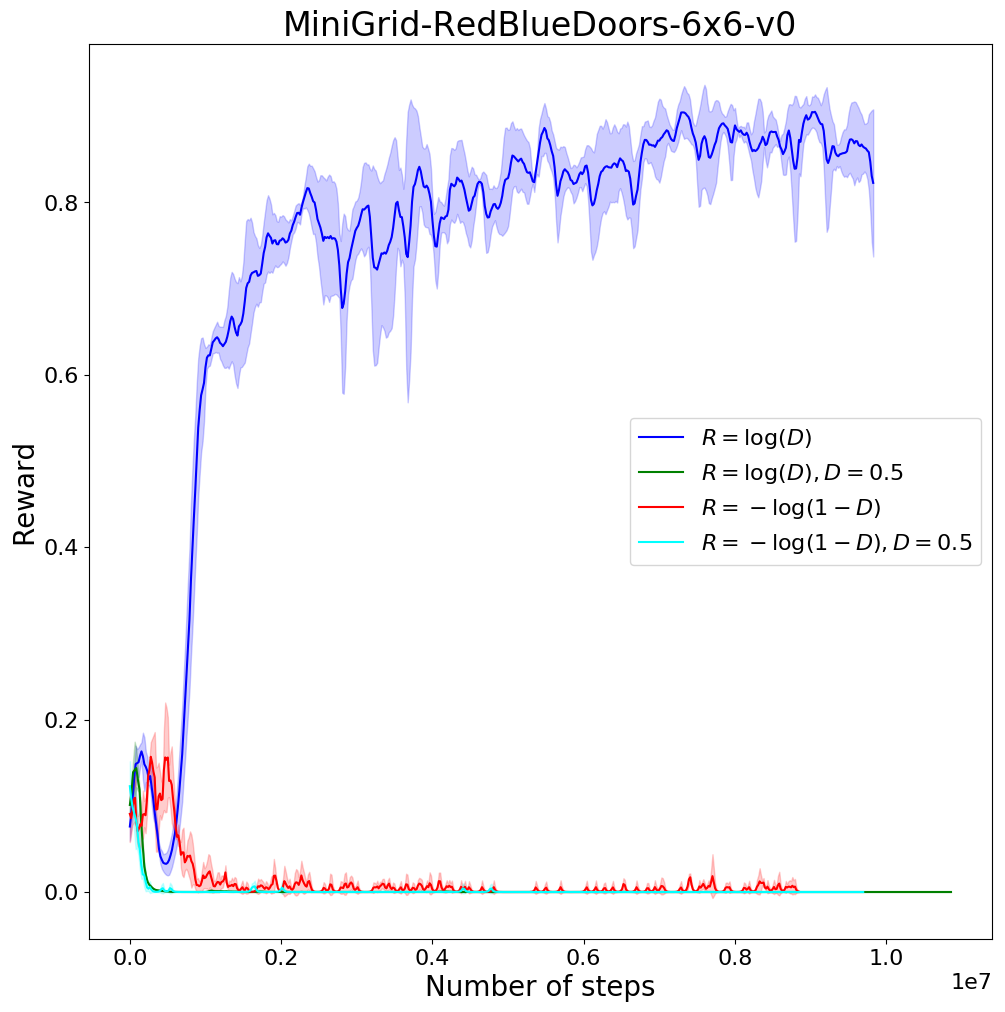}
    \caption{Episode rewards for different environments and different GAIL reward functions. The positive rewards show a survival bias because they do not complete the task (leading to low reward). The negative rewards show a termination bias, and in the simpler environments, the agent learns to complete the task even without trajectories ($D=0.5$) only due to the reward bias.}
    \label{fig:notrajrewards}
\end{figure}

%\begin{figure}
    %\centering
    %\includegraphics[width=0.32\textwidth]{notrajlengths/MiniGrid-Empty-Random-6x6-v0.png}
    %\includegraphics[width=0.32\textwidth]{notrajlengths/MiniGrid-DoorKey-6x6-v0.png}
    %\includegraphics[width=0.32\textwidth]{notrajlengths/MiniGrid-DistShift1-v0.png}
    %\includegraphics[width=0.32\textwidth]{notrajlengths/MiniGrid-GoToDoor-6x6-v0.png}
    %\includegraphics[width=0.32\textwidth]{notrajlengths/MiniGrid-RedBlueDoors-6x6-v0.png}
    %\caption{Episode lengths for different environments and different GAIL reward functions. The GAIL with positive reward functions always saturate at the maximum possible episode length, essentially defaulting to looping behavior.}
    %\label{fig:notrajlengths}
%\end{figure}

\subsubsection{Survival bias in positive rewards}
% In a task-based environment, the agent can potentially exploit the reward function by the discriminator.
In a task-based environment, providing non-negative rewards at every step may lead to survival bias and the agent may not complete the task and loop in the environment to collect more rewards.
% For example, if we assume that each expert state has at most one expert action, and there are states which the expert does not visit (for example, the grid cells in the other rooms).
To show that survival bias is indeed an issue, we consider an ideal case.
We consider an ``oracle discriminator'', which gives a positive reward $R$ when the agent performs the expert action for a given state, and 0 otherwise.
In other words, the reward function of the discriminator is $R(s,a) = R$ if $a = \pi_E(s)$ and $R(s,a) = 0$ if $a \ne \pi_E(s)$.
In practice, the discriminator reward is also bounded to prevent stability issues by using $R(s,a) = -\log(\max(\epsilon, 1 - D(s,a)))$ which has range $\Big[0, \log(\frac{1}{\epsilon})\Big]$.
Now consider an expert trajectory of path length $p$.
Since the discriminator will give a positive reward $R$ at each step in the expert trajectory, the total discounted reward of the expert trajectory  is:
\begin{align}
R_E &= R + \gamma R + \ldots \gamma^{(p-1)} R = \frac{1 - \gamma^p}{1 - \gamma}R
\end{align}
Now consider a trajectory where the agent follows the expert trajectory for $p-1$ steps, and then loops between an expert action and non-expert action to survive for as long as possible.
The total discounted reward for this trajectory is:
\begin{align}
    R_{loop} &= R + \gamma R + \ldots \gamma^{(p-2)} R + 0 + \gamma^{(p)} R + 0 + \gamma^{(p+2)} R \ldots \\
    &= \frac{1 - \gamma^{(p-1)}}{1-\gamma} R + \frac{\gamma^p}{1 - \gamma^2}R 
\end{align}
For the agent to prefer this trajectory over the expert trajectory, $R_{loop}$ should be greater than $R_E$.
This inequality will imply the following:
\begin{align}
    \frac{1 - \gamma^{(p-1)}}{1-\gamma} R + \frac{\gamma^p}{1 - \gamma^2}R &\ge \frac{1 - \gamma^p}{1 - \gamma}R \\
    \implies \frac{\gamma^p}{1 + \gamma} &\ge \gamma^{(p-1)} (1 - \gamma)
    \implies \gamma \ge 1 - \gamma^2 
\end{align}
which is true for $\gamma \ge 0.6180$. In practical scenarios, $\gamma \approx 1$, and hence, the agent can prefer to loop in the environment rather than following the expert trajectories. 
It is therefore evident as to why positive rewards might be unsuitable for task based environments.

% \begin{figure}
    % \centering
    % \includegraphics[width=0.49\textwidth]{arch.png}
    % \includegraphics[width=0.49\textwidth]{discr.png}
    % \caption{Neural network architectures used in all experiments. The number inside the modules are the parameters - number of output channels for convolutions and output nodes for Linear layers. The parameter for leaky relu is 0.1}
    % \label{fig:arch}
% \end{figure}

\subsubsection{Termination bias in Negative Rewards}
Positive reward functions are prone to survival bias, and the next obvious choice of a reward function is $R(s,a) = \log(D(s,a))$.
This reward function is always negative since $D(s,a) \in [0, 1]$.
However, this reward function makes an agent prone to termination bias.
In termination bias, the agent tends to end the episode as soon as possible to stop accruing more negative rewards.
This may be beneficial for task based environments where the agent wants to complete the task as soon as possible.
However, if there are faster ways to terminate an episode, the agent may be biased to do that instead of completing the task.
In our experiments, we show that GAIL with the reward function $R(s,a) = \log(D(s,a))$ fails to learn the task when there are multiple termination conditions in the environment.
% In case of survival based environments like Hopper, Cheetah etc, the negative function is not able to learn the expert policy. Since, at each step the reward received by the policy is negative regardless of the action taken by the policy, so the policy learns to terminate the episode early by dying and thereby receiving minimum negative. For this reason, negative rewards are unsuitable for survival based environments.

\subsection{Discriminator Actor Critic}
Both the survival and task based environments suffer from different problems but for the same reason - the reward for the terminal state is implicitly set to 0 in all these problems, which is why positive rewards become unsuitable for task based environments and negative rewards become unsuitable for survival based environments.
%The work in \cite{dac} attempts to solve this problem by augmenting the state with an additional dimension which indicates whether the given state is terminal. Also, the reward for the terminal state is learnt explicitly.

To address the issue of implicit bias in the reward function of the GAIL, \cite{dac} propose learning rewards for absorbing state explicitly. Consider a trajectory of T time steps, let the reward for the terminal state i.e the Tth step be $R_T = R(s_T, a_T)$, in the standard GAIL setting as in Equation \ref{eq:gail}. In DAC, a new reward function for terminal state is defined $R_T = R(s_T, a_T) + \Sigma_{T + 1} ^ \infty \gamma ^{t - T}R(s_a, .)$ where $R(s_a, .)$ is learnt.
% With this formulation, GAIL is able to learn to determine whether terminating an episode at a certain is good or not and therefore, is able to prevent premature termination of episodes in survival based environments.  
This learnt reward for the terminal state removes the bias towards avoiding or transitioning to the terminal state.
With this formulation, GAIL tries to match the state occupancy of the expert and agent for the terminal state as well, hence mitigating survivor bias.
However, this modifies a finite-horizon environment into an infinite-horizon environment, and does not consider the scenario where the agent may terminate an episode in a multitude of different ways.
There are a few other problems as well that the work does not address:
\begin{enumerate}
    %\item \textbf{No on-policy version}: The replay buffer can be easily modified in an off-policy way.
    %However, in on-policy DAC, the transitions from terminal state to itself has to be continued infinitely.
    %It can also be extended finitely which may affect the learnt value of the terminal state.
    \item \textbf{Requires modifying the environment}: In the original implementation of DAC, wrappers are provided to augment the environment with an extra terminal state.
    However, this may not always be feasible as we may only have an immutable API to the environment.
    This may also be a problem in real environments.
    \item \textbf{Does not handle multiple terminal states}: In a task-based environment with multiple terminal states, the agent can terminate an episode in multiple ways, out of which only some may correspond to task completion, and this can hinder learning.
    If a large portion of the expected reward comes from being in the terminal state, then the agent may not distinguish between expert trajectories and other similar trajectories which terminate the episode, as long as the state occupancy of the terminal states match for the agent and the expert. 
    \item \textbf{May be sample inefficient}: In the original DAC paper, the method is off-policy but it is compared with other on-policy algorithms, which is unfair.
    We suspected that DAC may be sample inefficient as compared to GAIL because DAC also has to match the state occupancy of the terminal state in addition to the existing states in the environment.
    For our experiments, we implement an on-policy version of DAC and observe that the method indeed is less sample efficient than GAIL.
\end{enumerate}

\subsection{Towards an unbiased reward function}
\iffalse
Positive rewards lead to `survival bias` in the environment, and the agent can never learn the task.
Negative rewards, i.e $log(D)$ lead to `termination bias` in the environment, which may lead to suboptimal behavior when other termination conditions are provided. 
\fi
We introduce the notation of "neutral reward functions" to denote reward functions that are real valued.
For example, the reward function $R(D) = \log(D) - \log(1-D)$ has the range $(-\infty, \infty)$ for $D \in (0, 1)$ and is real-valued.
Neutral reward functions can be unbiased because they can penalize looping behavior with negative rewards, effectively ``cancelling'' the positive reward that they acquired from previous loops (assuming that the expert doesn't loop itself).
The reward function can also potentially overcome `termination bias' because the agent can collect as many positive rewards as possible by following the expert trajectories.
To give an intuition of these claims, we show a similar theoretical sketch as before. 

Consider an oracle discriminator that gives reward $R$ when the agent takes the expert action $a$ to a state $s$, and gives a negative reward $-R$ otherwise (since the neutral reward function is a symmetric function $R(s, a) = \log(D(s,a)) -\log(1 - D(s,a))$ ). 
To ensure the agent learns to perform the task, the discriminator should assign the highest rewards to the expert trajectories.
Consider an expert trajectory of length $p$.
Similar to the previous example, assume that each state has only one expert action.
The reward of the expert trajectory is:
\begin{align}
R_E = R + \gamma R + \ldots \gamma^{(p-1)}R = \frac{1 - \gamma^p}{1 - \gamma} R \label{eq:re_neutral}
\end{align}
The maximum reward for any other trajectory can be obtained by mimicking the expert till timestep $p-1$ to get positive rewards and do a non-expert action (to prevent the trajectory from becoming an expert trajectory).
To recover from the negative reward just incurred, the agent performs an expert action.
After this, it cannot perform an expert action again because that will complete the expert trajectory and the episode will terminate.
So, if it wants to loop it will perform a non-expert action followed by an expert action and continue doing so.
Hence the reward for this trajectory $loop$ is:
\begin{align}
    R_{loop} &= R + \gamma R + \ldots \gamma^{(p-2)}R - \gamma^{(p-1)} R + \gamma^{p}R - \gamma^{p+1}R \ldots \\
    &= \frac{1 - \gamma^{(p-1)}}{1 - \gamma}R -  \frac{\gamma^{(p-1)}}{1 + \gamma}R
\end{align}
For an agent to prefer surviving rather than completing the expert trajectory,  we need to have $R_{loop} \ge R_E$. This implies:
\begin{align}
    \frac{1 - \gamma^{(p-1)}}{1 - \gamma}R -  \frac{\gamma^{(p-1)}}{1 + \gamma}R &\ge \frac{1 - \gamma^p}{1 - \gamma} R \\
    \implies \frac{\gamma^{(p-1)} - \gamma^p}{1 - \gamma} \le - \frac{\gamma^{(p-1)}}{1 + \gamma} \implies 1 \le \frac{-1}{1 + \gamma}
\end{align}
Which is impossible for a discounting factor $\gamma \in [0, 1]$. Hence, $R_{loop}$ will always be lesser than $R_E$ for an oracle discriminator, and the agent will always prefer following the expert trajectory.

A similar argument can be made for neutral rewards overcoming termination bias. 
Consider the previous setup, but the agent tries to terminate the episode before completing the task.
The expert reward is already in Equation \ref{eq:re_neutral}.
To exhibit termination bias, the agent has to terminate the episode in at most $p$ steps, and the last action should be a non-expert action (if its an expert action, then the agent is exactly imitating the expert).
Hence, the maximum reward the agent can accrue is :
\begin{align}
    R_{term} = R + \gamma R + \ldots \gamma^{(p-2)}R - \gamma^{(p-1)}R
\end{align}
which is the same as $R_E$ for the first $p-1$ terms but the last term is a negative term whereas the last term of $R_E$ is a positive term.
Since $R_E \ge R_{term}$, the agent will prefer the expert trajectory over some other shorter non-expert path.
Hence, the neutral reward overcomes termination bias.

\subsection{Choice of Environment}

\begin{wrapfigure}{l}{0.4\textwidth}\centering
%\begin{minipage}[t]{0.4\textwidth}
%\begin{figure}
    %\centering
    \includegraphics[width=0.19\textwidth]{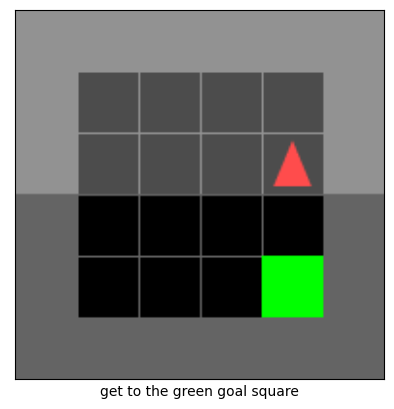}
    \includegraphics[width=0.19\textwidth]{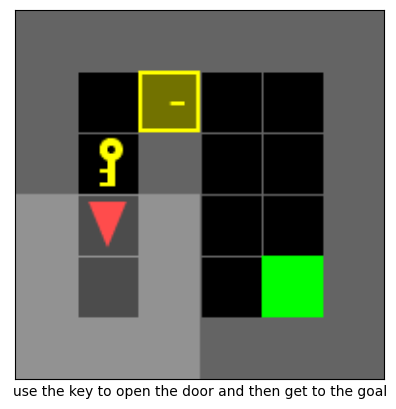}
    \includegraphics[width=0.19\textwidth]{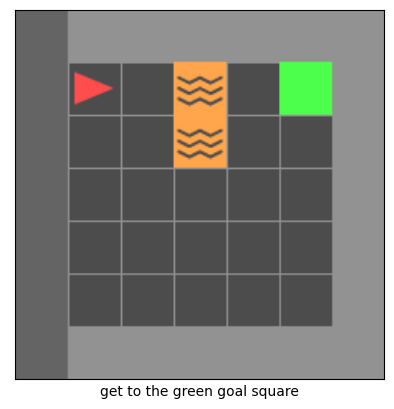}
    \includegraphics[width=0.19\textwidth]{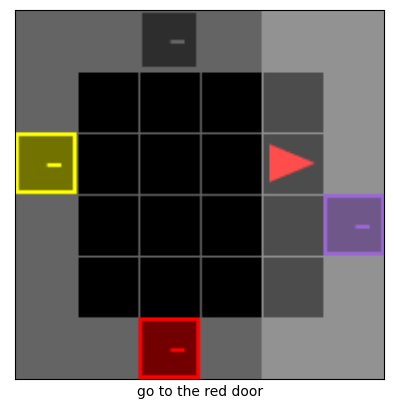}
    \includegraphics[width=0.39\textwidth]{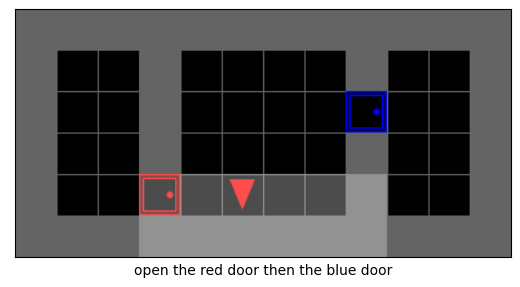}
    \caption{Environments used in the paper. The top-left and top-middle environments terminate only on reaching the goal (green). The other environments can terminate through other conditions as well (mentioned in the paper). However, expert trajectories consist of goal-directed behaviors only.}
    \label{fig:envs}
    \vspace*{-38pt}
%\end{figure}
%\end{minipage}
\end{wrapfigure}
To observe the limitations of the current methods used in adversarial imitation learning, we propose to use task based environments. 
However, task based environments with single terminal states will not be able to test the bias of the reward functions due to termination conditions.
Hence, we propose to use environments from the Gym Minigrid package \cite{minigrid} with single and multiple termination conditions.

We disable the \textit{Done} action to examine the bias of the agent only due to environmental termination conditions.
To verify the effects of reward bias in a single termination condition, we choose the \textit{Empty} and \textit{DoorKey} environments.
\begin{enumerate}
    \item In the \textit{Empty} environment, the agent starts at a random location in the grid and the episode terminates when the agent steps on the goal location (in green).
    \item In the \textit{DoorKey} environment, the agent starts in a room and has to pick up a key to open a locked door. The agent has to reach the goal on the other side of the locked door.
\end{enumerate}
Both these environments have only one termination condition - the agent has to reach the goal location.
Methods like Discriminator-Actor-Critic \cite{dac} can learn a non-zero reward for the terminal condition which overcomes the survival bias of the positive reward function.
GAIL with a negative reward can also learn to solve these tasks since they want to accumulate the maximum reward which would require termination of the episode as soon as possible.

However, both of these methods can suffer from the same problem - multiple termination conditions.
The negative reward will have a bias towards trajectories that terminate early to accumulate fewer negative rewards.
The DAC baseline will not differentiate between different termination conditions (termination due to goal completion v/s termination due to other conditions) since the terminal state can potentially provide a large cumulative reward.
To test this, we use three other environments within Gym Minigrid. They are:

\begin{enumerate}
    \item \textit{RedBlueDoors}: In this environment, the task is to open a red door, followed by opening the blue door.
    The episode terminates when the blue door is opened regardless of whether the red door was opened or not.
    \item \textit{GoToDoor}: This environment is similar to the \textit{RedBlueDoors} environment but there are 4 doors of different colors, and the episode terminates when any door is opened.
    However, the task is considered to be completed only if the red door is opened.
    \item \textit{DistShiftv0}: In this environment, the agent has to cross a room with lava near the walls of the room (refer to Figure \ref{fig:envs}.
    The task is to reach the goal location and avoid the lava.
    The episode ends when the agent touches the lava or goal.
\end{enumerate}
These environments have termination conditions which will affect the variants of GAIL depending on the reward bias that they may have.
We hypothesize that the inability of DAC to differentiate between terminal states would lead to suboptimal performance in each of these environments.

%===============================================================================

\begin{figure}
    \centering
    \includegraphics[width=0.32\textwidth]{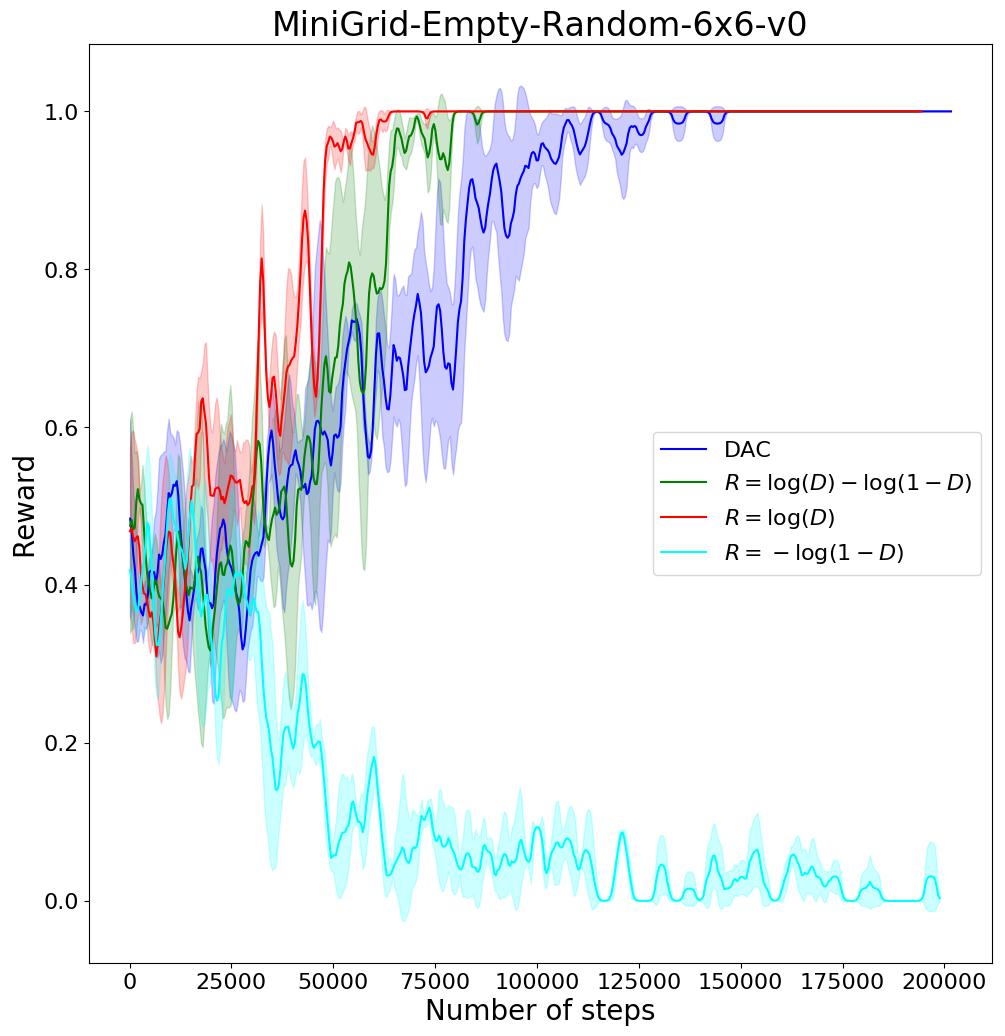}
    \includegraphics[width=0.32\textwidth]{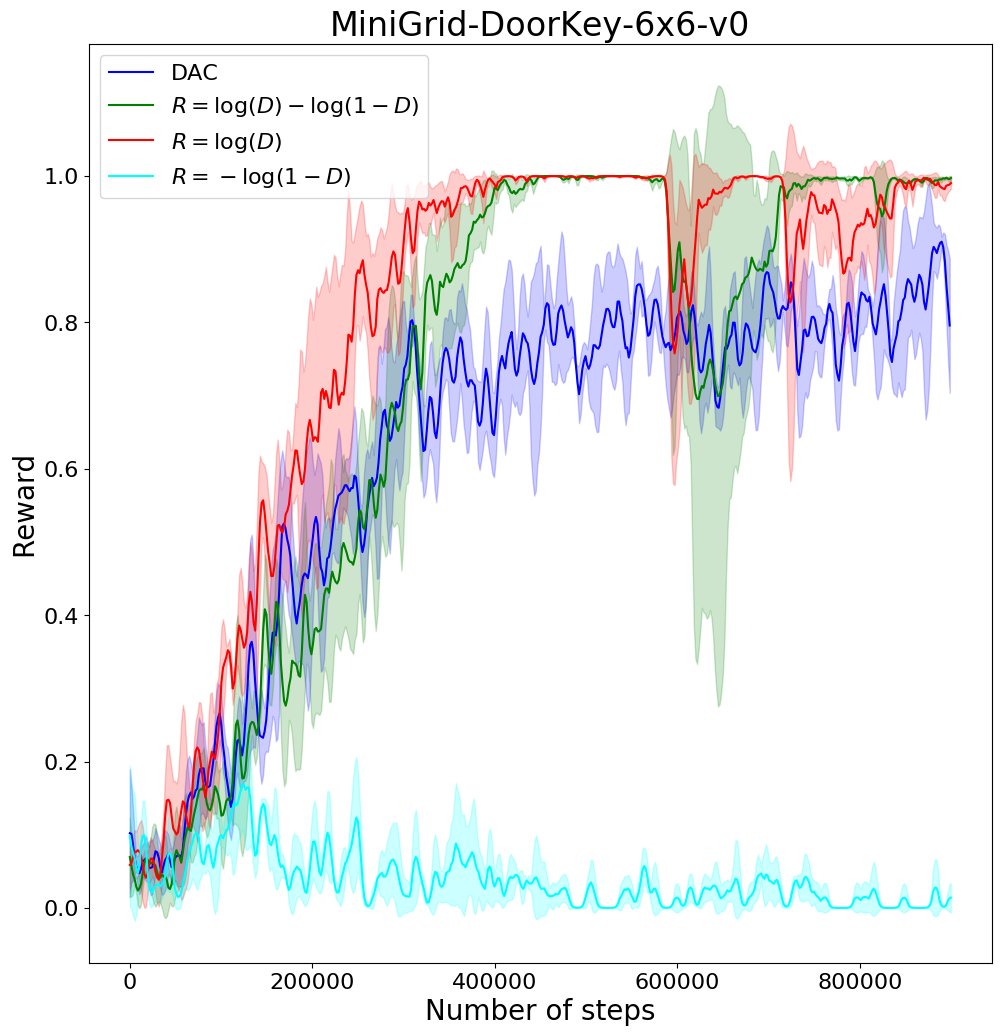}
    \includegraphics[width=0.32\textwidth]{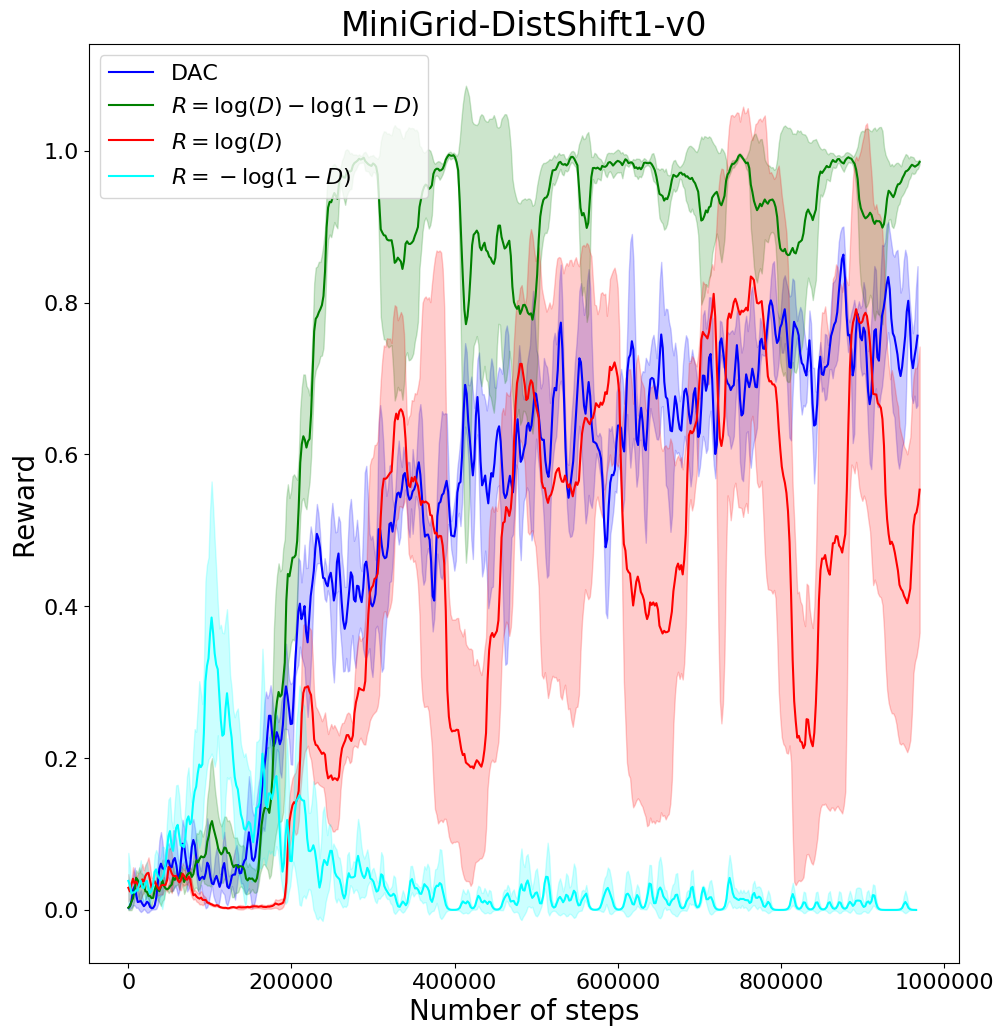}
    \includegraphics[width=0.32\textwidth]{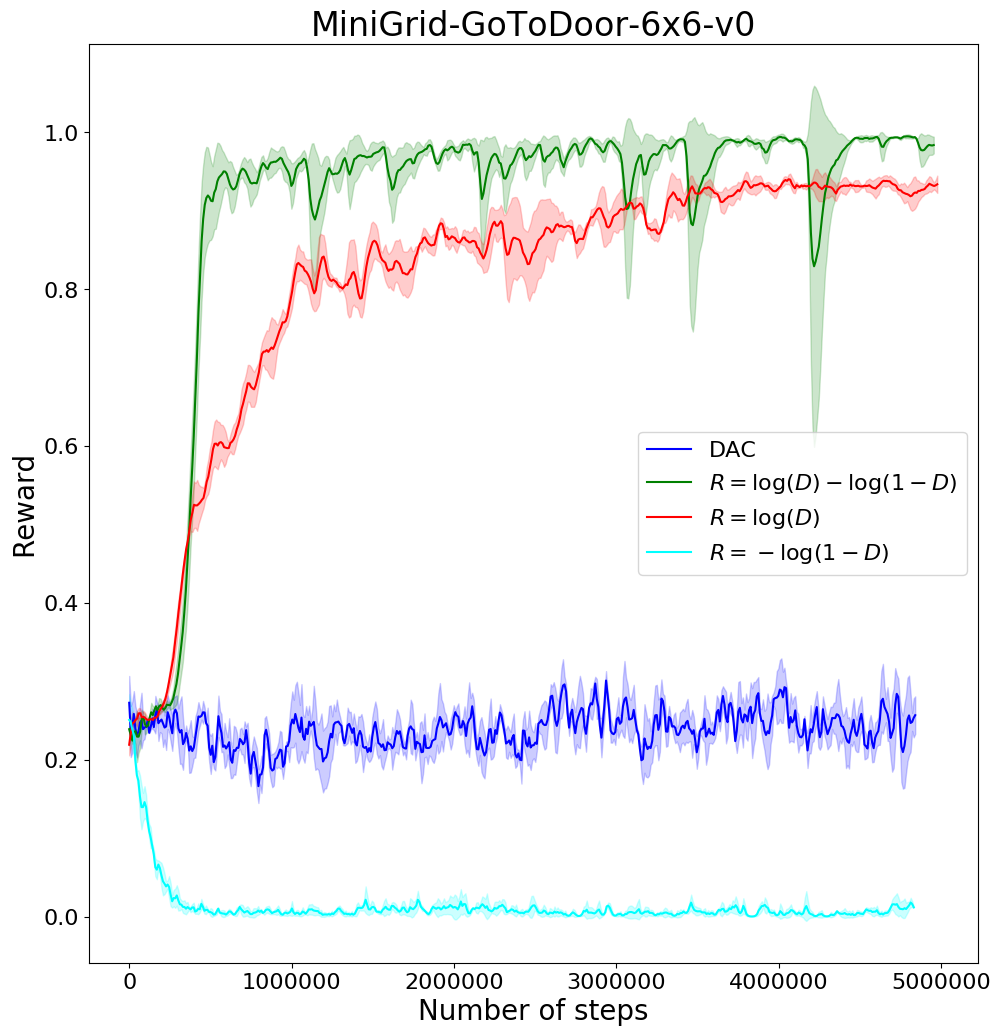}
    \includegraphics[width=0.32\textwidth]{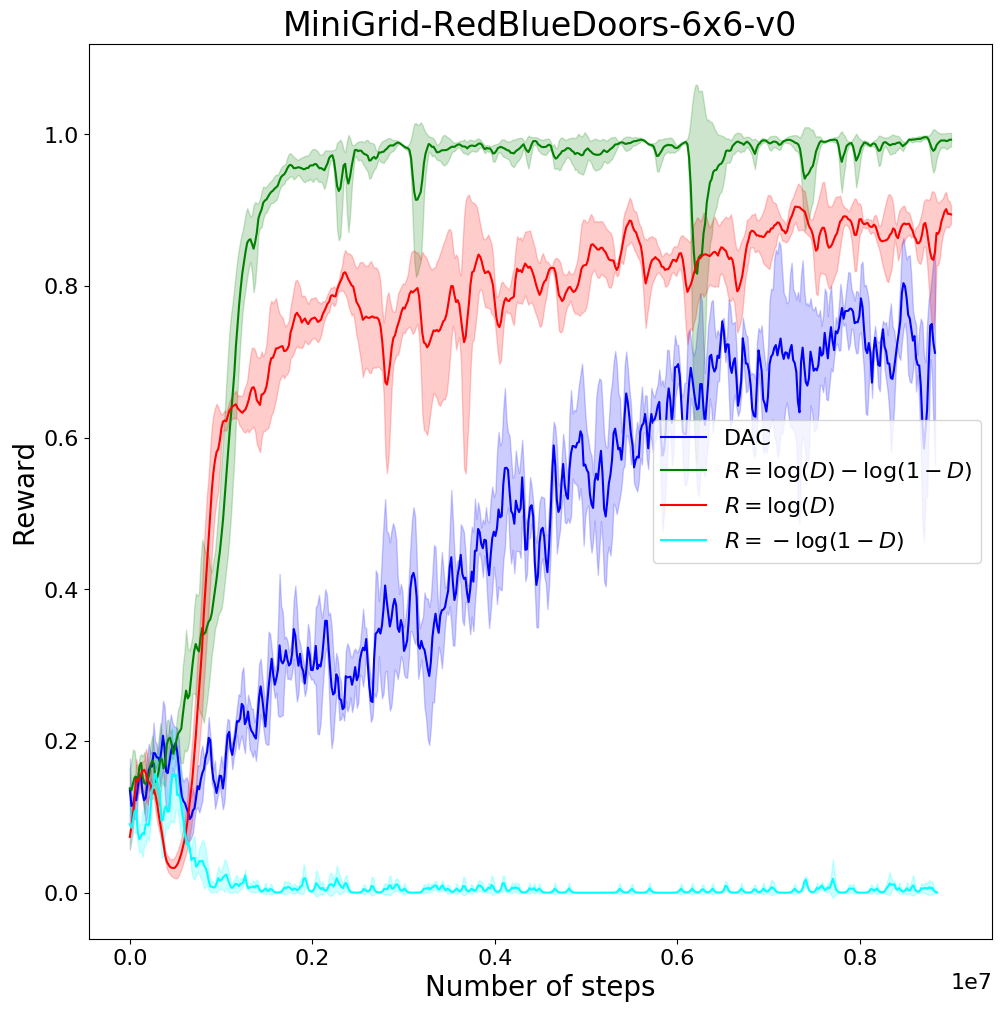}
    \caption{Comparison of performance of different GAIL variants. On-policy DAC is slow in terms of converging to the optimal policy, possibly due to unstable learning of the terminal state reward. In environments with non-goal terminal states, DAC and GAIL with negative rewards converge to suboptimal policies due to their respective biases.}
    \label{fig:rewards}
\end{figure}

\section{Experiments}
\vspace{-5pt}
Our expert is a trained PPO policy. We use 1000 rollouts from the expert policy to train our imitation learning agents for all our experiments.  
Each rollout has $\sim$10 state-action pairs, so the total number of state action pairs are comparable to \cite{babyai} and \cite{gail}.
The locations and colors of the objects are also perturbed across episodes, so the agent has to generalize to the concept of objects rather than memorizing the goal locations.
We use 3 CNN layers with Leaky ReLU activations to extract a state embedding of the environment as shown in Figure \ref{fig:arch} for all our experiments.
We use PPO \cite{ppo} for on-policy optimisation of the GAIL reward function for all experiments.
In each case, the policy network uses this state and returns probability of an action given a state. 
The policy network head is a 2-layered MLP with Leaky ReLU activations.
The value network head is identical to the policy network as shown in Figure \ref{fig:arch}.
We average all our results across 3 random seeds.
The objective of our first experiment is to exhibit the implicit bias in the positive and negative reward functions. We compare the environment rewards achieved by the agents and the success rate for the 5 environments that we had mentioned earlier. 
For our second experiment, we compare the environment rewards and success rate of (i) DAC \cite{dac} (ii) Positive Reward (iii) Negative Reward (iv) Neutral Rewards in the five environments.
\iffalse
\begin{enumerate}
    \item \textbf{Positive Reward}: Uses $-log(1 - D)$ as the reward. In this we evaluate two scenarios one where D is being learnt and the other where D is set as a constant equal to 0.5.
    \item \textbf{Negative Reward}: Uses $log(D)$ as the reward.  In this we evaluate two scenarios one where D is being learnt and the other where D is set as a constant equal to 0.5.
    \item \textbf{Neutral Reward}: Uses $log(D) - log(1 - D)$ as the reward.
    \item \textbf{DAC}: Learn reward for the absorbing state as proposed by Kostrikov et al\cite{dac}. 
\end{enumerate}
\fi
\section{Results}
\vspace{-5pt}
Figure \ref{fig:notrajrewards} shows reward curves for positive and negative rewards with and without learning the discriminator.
The first two environments, \textit{Empty} and \textit{Doorkey} are able to attain expert like performance with negative rewards, which was expected as these are task based environments with single termination state.
GAIL with positive reward is not able to learn the expert policy due to its survival bias (first two plots of the Figure \ref{fig:notrajrewards}).
In case of \textit{Empty}, the agent is able to learn the task even without training the discriminator which shows that the negative reward function has a termination bias strong enough for task based environments.
Similarly, for the \textit{DoorKey} environment, GAIL with negative reward starts to learn the task better than positive rewards even when the discriminator is not trained.
However, in last three plots of Figure \ref{fig:notrajrewards}, where environments with multiple terminal states are used, termination bias is not enough to learn the task.
Positive rewards fall into the survival bias and don't learn the task.
An interesting observation is that the success rate of \textit{GoToDoor} with negative reward and no trajectories is 0.25 as shown in Figure \ref{fig:notrajrewards}.
The termination bias makes the agent reach out to the nearest door to end the episode, which has a 25\% rate of being the red door.
\begin{table}[htpb]
    \centering
    \begin{adjustwidth}{-1cm}{}
    \resizebox{1.1\textwidth}{!}{
    \centering
    \begin{tabular}{|l|c|c|c|c|c|} \hline
     & \textbf{Empty} & \textbf{DoorKey} & \textbf{GoToDoor} & \textbf{RedBlueDoors} & \textbf{Distshift1} \\ \hline
    $R = -\log(1-D)$ & $0.03 \pm 0.15$ & $0.00\pm 0.00$ & $0.00\pm 0.00$ & $0.00\pm 0.00$ & $0.00\pm 0.00$ \\ \hline
    $R = -\log(0.5)$ & $0.00\pm 0.00$ & $0.00\pm 0.00$ & $0.00\pm 0.00$ & $0.00\pm 0.00$ & $0.00\pm 0.00$\\ \hline
    $R = \log(D)$ & $\boldsymbol{1.00 \pm 0.00}$ & $\boldsymbol{1.00\pm 0.00}$ & $0.93 \pm 0.24$ & $0.83 \pm 0.37$ & $0.85 \pm 0.35$\\ \hline
    $R = \log(0.5)$ & $\boldsymbol{1.00 \pm 0.00}$ & 0.25 $\pm$ 0.43 & 0.38 $\pm$ 0.48 & 0.00 $\pm$ 0.00 & 0.65 $\pm$ 0.47\\ \hline
    DAC & $\boldsymbol{1.00 \pm 0.00}$ & 0.83 $\pm$ 0.37 & 0.26 $\pm$ 0.44 & 0.70 $\pm$ 0.45 & 0.76 $\pm$ 0.42 \\ \hline
    $R = \log(D) - \log(1-D)$ & {$\boldsymbol{1.00 \pm 0.00}$} & {$\boldsymbol{1.00 \pm 0.00}$} & {$\boldsymbol{0.97 \pm 0.15}$} & {$\boldsymbol{0.91 \pm 0.27}$} & {$\boldsymbol{0.97 \pm 0.15}$} \\ \hline
    \end{tabular}
    }
    \caption{Success rate of all learnt policies across 3 seeds. For each seed, 20 trajectories were recorded and the mean and variance is calculated across all trajectories.}
    \label{tab:perf}
    \end{adjustwidth}
\end{table}
In Figure \ref{fig:rewards}, we compare the performance of DAC, neutral rewards, negative and positive rewards.
The success rate of the different methods is also summarized in Table \ref{tab:perf}.
For \textit{Empty} and \textit{DoorKey} all methods except GAIL with positive rewards are able to learn the task.
However, DAC is slower than other GAIL methods, and we hypothesize this is because DAC has to learn the reward for the terminal state as well.
%we observe, all three, DAC, neutral rewards and negative rewards are able to learn the expert policy well enough. However, DAC is slower than neutral and negative rewards to converge to the expert policy. The positive reward again performs poorly due its inherent survival bias. 
%For the other 3 environments, negative reward isn't able to attain the reward of 1 and hence is not able to learn the expert policy well enough.
For the other three environments, termination bias hinders learning with negative rewards and DAC suffers in performance due to lack of distinction between terminal states.
For \textit{GoToDoor}, \textit{DistShift} and \textit{RedBlueDoors}, DAC is not able to achieve high success rate showing its limitations in task based scenarios with multiple terminal states. 
Success rate of neutral reward is the highest and better than the negative reward showing its robustness to termination bias.
Similarly, for other the other two task based environments with multiple terminal states, neutral reward significantly outperforms both DAC and negative reward.
GAIL with neutral rewards is able to outperform all other methods, hence, is able to overcome both survival and termination bias.
%DAC is again slow to converge is not able to achieve expert performance.
%And positive reward again fails due to its survival bias.
%However, the neutral reward is able to replicate the expert policy and attain the reward of 1 in all these 3 environments as well.
%As we can observe from the table, for Empty all the methods except for positive reward is able to achieve high success rate, which was expected since DAC, negative rewards and neutral rewards all work well with task based environments with multiple terminal states.
%Even, negative reward without training the discriminator is able to achieve expert performance pointing to the high reward bias.
%Similarly, in doorkey, all methods except for positive rewards achieve high success rate. DAC\cite{dac} slower to converge, therefore, its performance is slightly lower. 
\section{Conclusion}
\vspace{-5pt}
In this work, we address the problem of reward bias in adversarial imitation learning.
We explore two types of reward biases - survival bias and termination bias, and how different reward functions lead to these biases in the agent.
Positive reward functions encourage survival bias, and negative reward functions encourage termination bias, and both these biases may hinder learning in a task-based environment.
We show that real-valued reward functions are unbiased and can learn to overcome both survival and termination biases.
Experiments show that this simple change of reward function enables the agent to imitate the expert on tasks with single and multiple termination conditions.

\iffalse
From our experiments, we are able to show that positive reward is not able to work well in task based environments.
Negative rewards are able to learn the expert policy in task based environments with single terminal state, however, they are unable to attain expert performance in task based environments with multiple terminal state.
DAC is slow to converge is not able to learn the expert policy in task based environments with multiple terminal states.
Our proposed method using neutral rewards achieves expert performance in task based environments with single termination state as well multiple termination states. 
\fi

\clearpage
\bibliography{example}  % .bib

\clearpage
% Supplementary material
\section{Supplementary Material}
\subsection{Extending to real environments}
We test our method in different simulation environments to check its robustness. It performs well in all the settings that we consider in our experiments. Further, the environments were carefully chosen keeping in mind the real life scenarios and possible practical applications. For instance, in a real life task based scenario such as a robot trying to navigate from a given point to another, the robot might have to also learn to avoid obstacles in the environment, i.e to avoid attaining certain states so as to survive in the environment as bumping into them might lead to irreversible damage to the robot. So, these might be regarded as real life task based scenarios with multiple terminal states. Simulated environments can be really useful in these settings as the agent can safely learn to avoid these bad terminal states and learn a `safe' policy which the agent can further fine tune in the real world by interacting with the environment. The environments: DistShift, RedBlueDoors and GotoDoor are environments which emulate this setting of multiple terminal states. We extend a standard repository \cite{code} for all experiments.

\subsection{Effect of different reward functions on episode length}
\begin{figure}[htpb]
    \centering
    \includegraphics[width=0.30\textwidth]{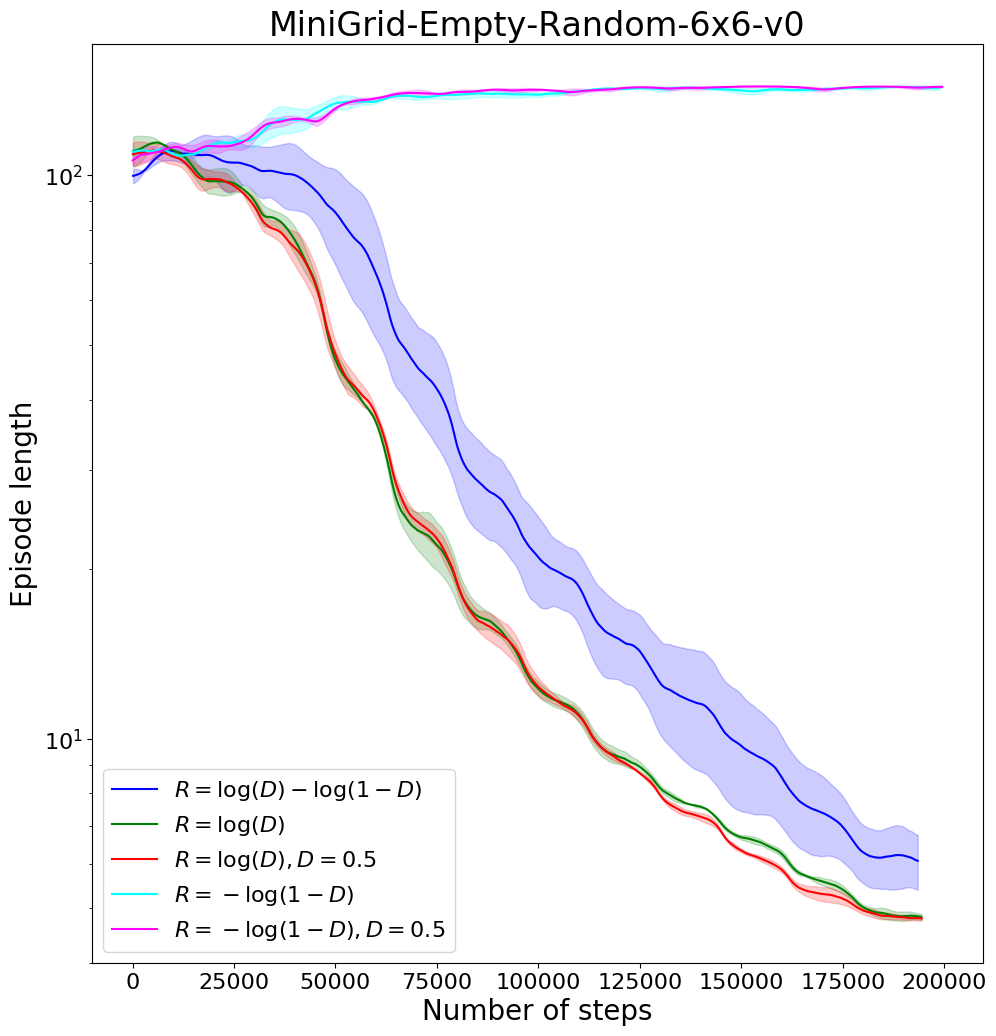}
    \includegraphics[width=0.30\textwidth]{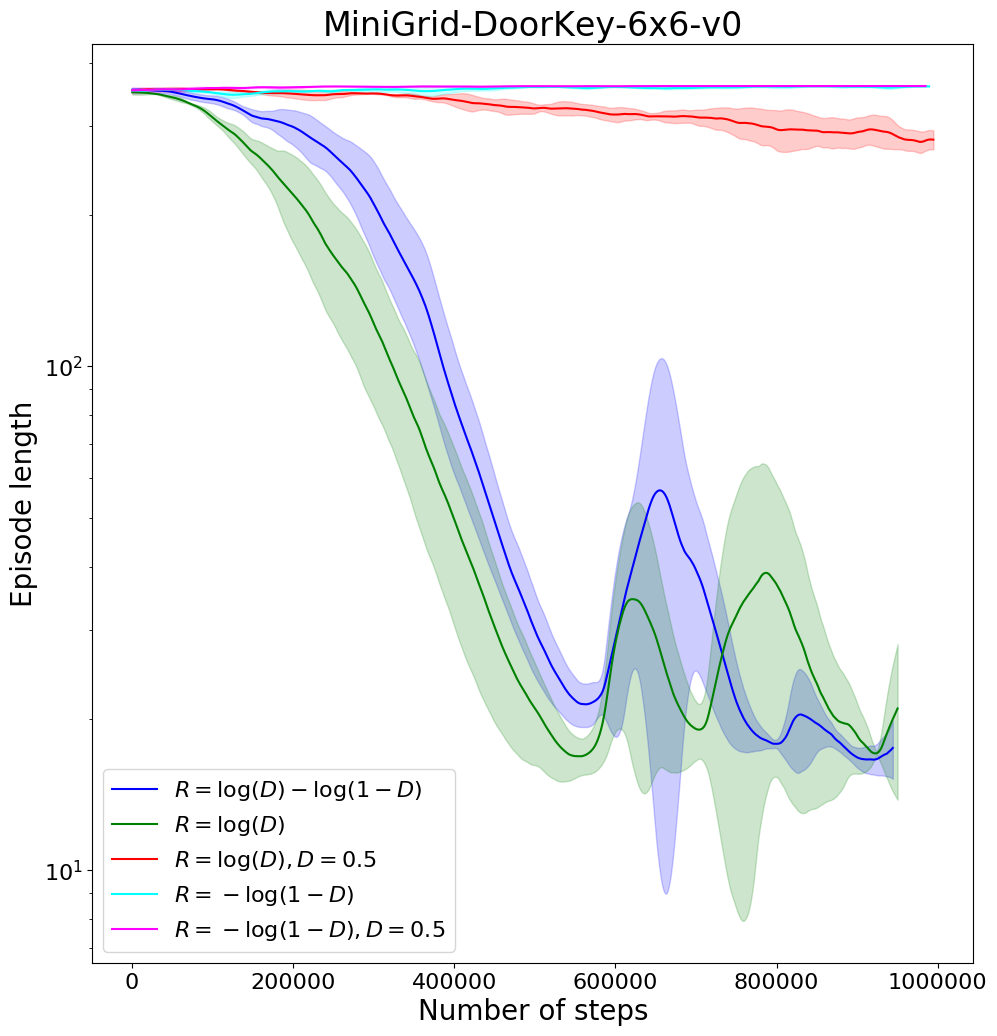}
    \includegraphics[width=0.30\textwidth]{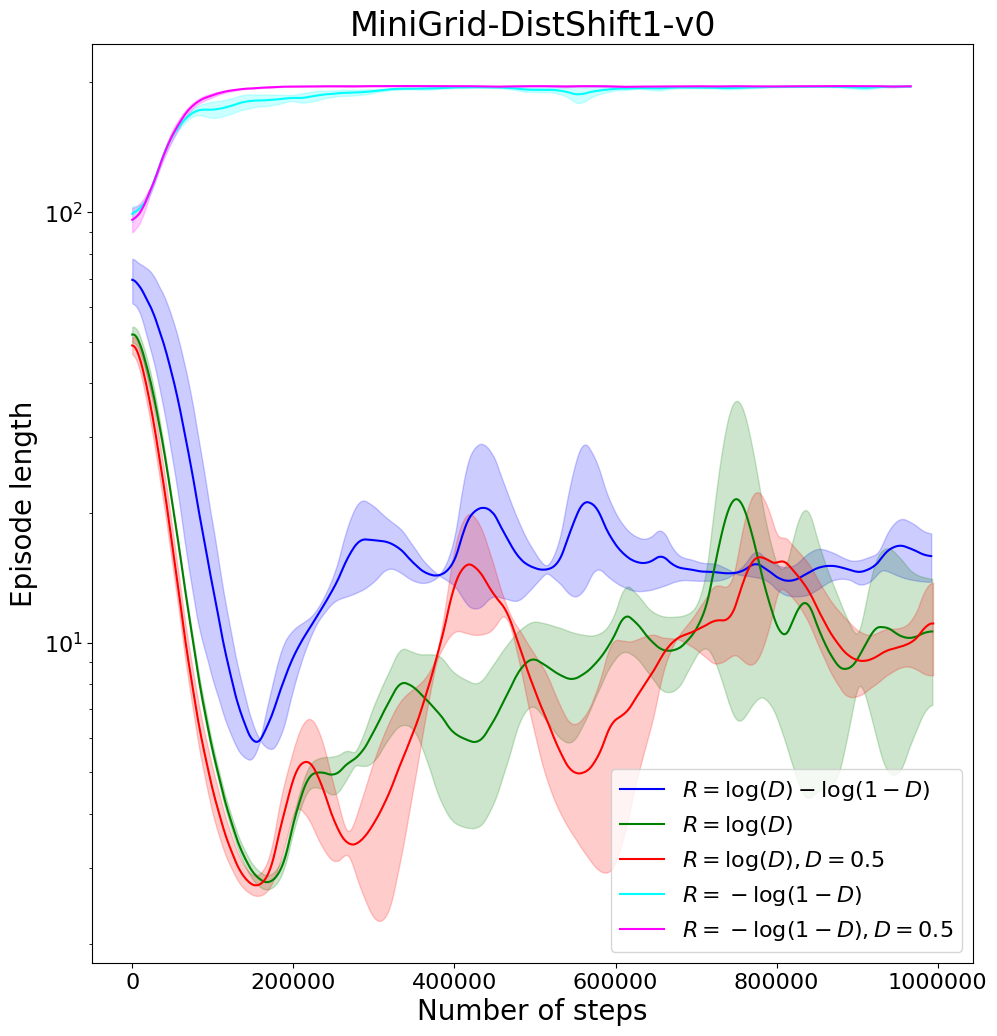}
    \includegraphics[width=0.30\textwidth]{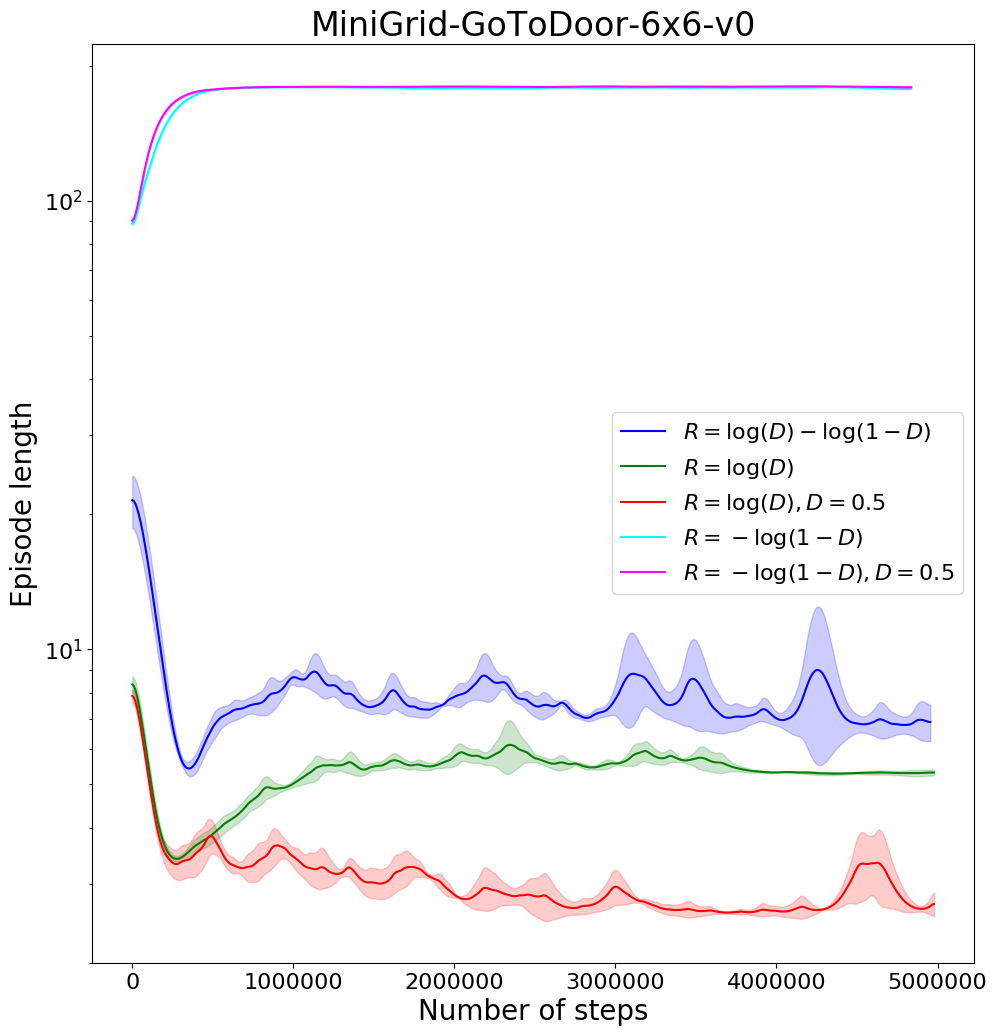}
    \includegraphics[width=0.30\textwidth]{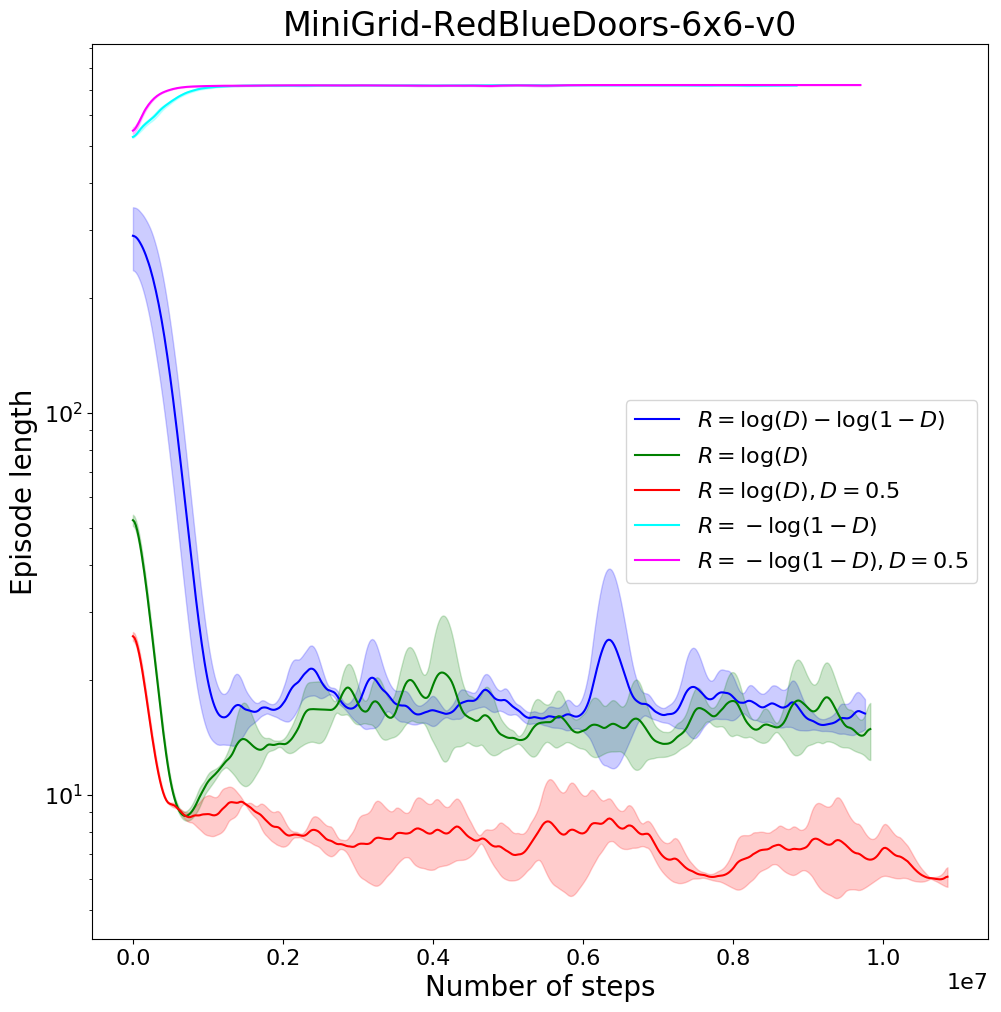}
    \caption{Comparison of average episode length for different GAIL reward functions.}
    \label{fig:lengths}
\end{figure}

In Figure \ref{fig:rewards} we analysed the effect of different GAIL rewards on the agent's ability to learn the task. 
Here, we confirm the effects of our method by checking the average episode lengths of the methods (in Figure \ref{fig:lengths}).
In all environments, GAIL with positive rewards fails to learn the task, and the episode lengths always reach the time-limit, therefore confirming that survival bias in GAIL is indeed a practical problem.
In environments like \textit{Empty} and \textit{DoorKey}, negative rewards are better since there is a single termination condition corresponding to task completion.
We see that this method indeed performs better than GAIL with positive rewards.
This reward is closely followed by neutral rewards which also obtain a similar curve for average episode length.
However, in the other environments with multiple terminal conditions, we can see a clear difference between negative and neutral rewards.
The difference is clearly visible in \textit{GoToDoor} where there is a non-negligible gap between negative and neutral rewards.
Although the average episode length for negative rewards is lower, the success rate is also lower, which confirms that negative rewards suffer from a termination bias problem. 
Neutral rewards take slightly more time on average to end an episode, but that is because they actually complete the task and do not terminate an episode early.

\end{document}